\documentclass{article}

\usepackage[preprint,nonatbib]{neurips_2021}

\usepackage[utf8]{inputenc} % allow utf-8 input
\usepackage[T1]{fontenc}    % use 8-bit T1 fonts
\usepackage{hyperref}       % hyperlinks
\usepackage{url}            % simple URL typesetting
\usepackage{booktabs}       % professional-quality tables
\usepackage{amsfonts}       % blackboard math symbols
\usepackage{nicefrac}       % compact symbols for 1/2, etc.
\usepackage{microtype}      % microtypography
\usepackage{xcolor}         % colors
\usepackage{graphicx}
\usepackage{multirow}
\usepackage{algorithm}
\usepackage{algorithmic}
\usepackage{subfigure}
\usepackage{amsmath}
\usepackage{amssymb}
\usepackage{amsthm}
\usepackage[page]{appendix}

\title{Fine-Grained AutoAugmentation for Multi-Label Classification}

\author{%
  Ya Wang\thanks{This work was completed during Ya Wang's internship at Alibaba DAMO Academy.}\\
  Peking University, Alibaba Group\\
  \And
   Hesen Chen,~
   Fangyi Zhang,~ 
   Yaohua Wang,~
   Xiuyu Sun\thanks{Corresponding author, E-mail: xiuyu.sxy@alibaba-inc.com.},~
   Ming Lin,~
   Hao Li \\
   Alibaba Group\\
   \texttt{xiuyu.sxy@alibaba-inc.com}\\
}

\begin{document}

\maketitle

\begin{abstract}
Data augmentation is a commonly used approach to improving the generalization of deep learning models. Recent works show that learned data augmentation policies can achieve better generalization than hand-crafted ones. However, most of these works use unified augmentation policies for all samples in a dataset, which is observed not necessarily beneficial for all labels in multi-label classification tasks, i.e., some policies may have negative impacts on some labels while benefitting the others. To tackle this problem, we propose a novel Label-Based AutoAugmentation (LB-Aug) method for multi-label scenarios, where augmentation policies are generated with respect to labels by an augmentation-policy network. 
%   learned via reinforcement learning. 
The policies are learned via reinforcement learning using policy gradient methods, providing a mapping from instance labels to their optimal augmentation policies. 
Numerical experiments show that our LB-Aug outperforms previous state-of-the-art augmentation methods by large margins in multiple benchmarks on image and video classification.
\end{abstract}

\section{Introduction}

Data augmentation is a powerful technique to enlarge a training dataset with more diversity. For a long time, augmentation policies are manually designed and achieve great success in various tasks, including object recognition \cite{real2019regularized}, image retrieval \cite{huang2020probability} and video activity recognition \cite{wang2020multi}. However, the manually designed augmentation policies heavily rely on individual experience and therefore suffer from individual bias. The transfer ability of these policies is also limited. Re-design of augmentation policies is normally required when facing new tasks, which could be time-consuming.
% The policy transfer to new tasks can also be costly, as augmentation policies for a new task can it could take a long time to design a good augmentation policy for this specific task since the transfer-ability of the manually designed policy is very limited. 
To remedy these drawbacks, approaches are recently proposed to automatically learn augmentation policies in a data-driven way~\cite{cubuk2019autoaugment,lim2019fast,zhang2019adversarial}, obtaining remarkable gains over manually designed ones.
% The learned policy consists of a set of pre-defined basic transformations which are jointly applied to the training instances with learned hyper-parameters. 
% These learned policies show remarkable gains over manually designed ones. 

% \begin{figure}[!t]
% 	\centering
% 	\includegraphics[width=\columnwidth]{./figures/label_aug.pdf}
% 	\caption{The gain/drop of augmentations on different labels of Peta. ``1'' indicates positive influence to the classification. ``-1'' presents the influence is negative and ``0'' means inconspicuous influence.}
% 	\label{fig:influence}
% \end{figure}

\begin{figure}[!t]
	\centering
	\subfigure[Gain/drop of augmentation]
  {\includegraphics[width=0.43\textwidth]{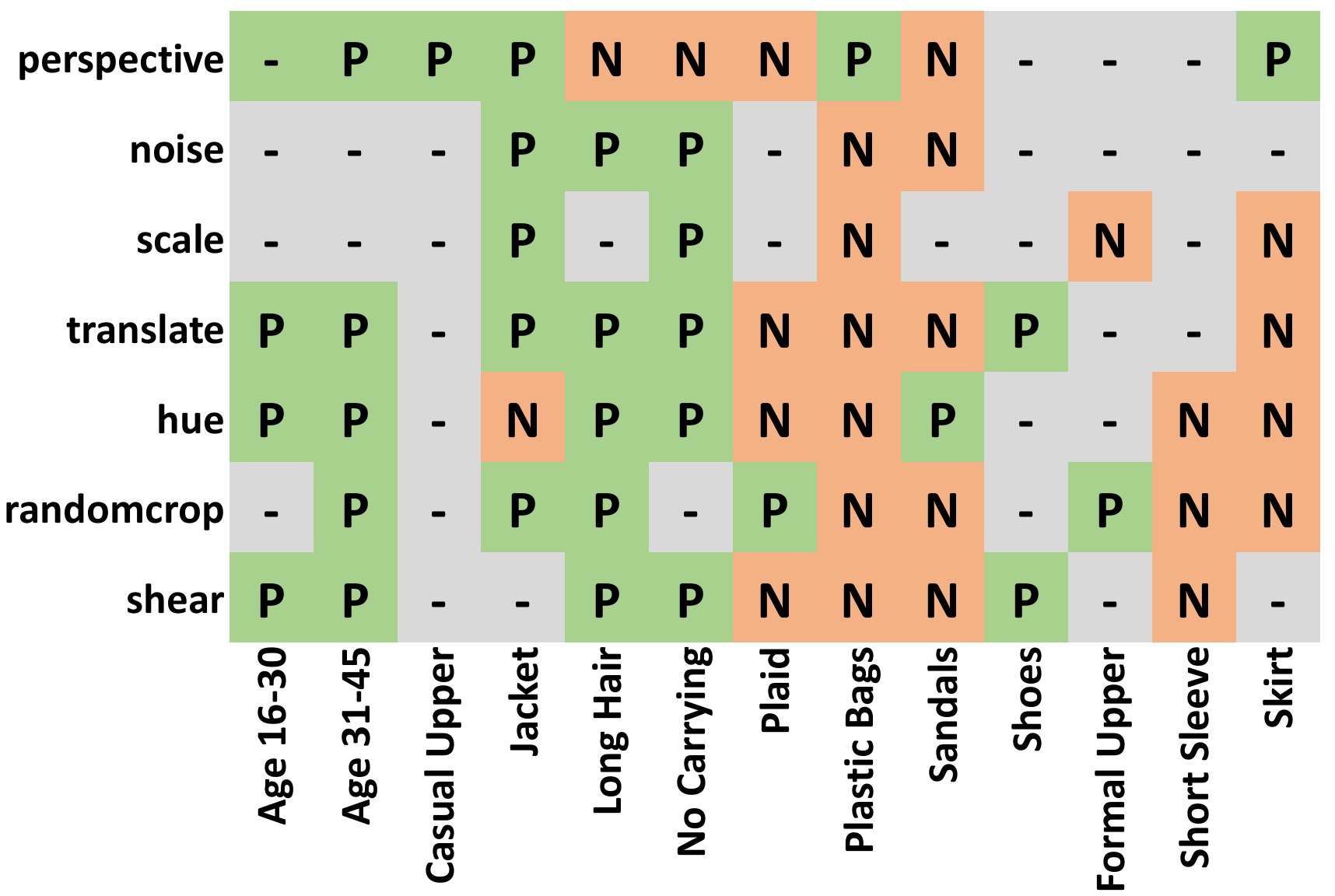} \label{fig:influence}} \quad
	\subfigure[Idea of LB-Aug] 
  {\includegraphics[width=0.49\textwidth]{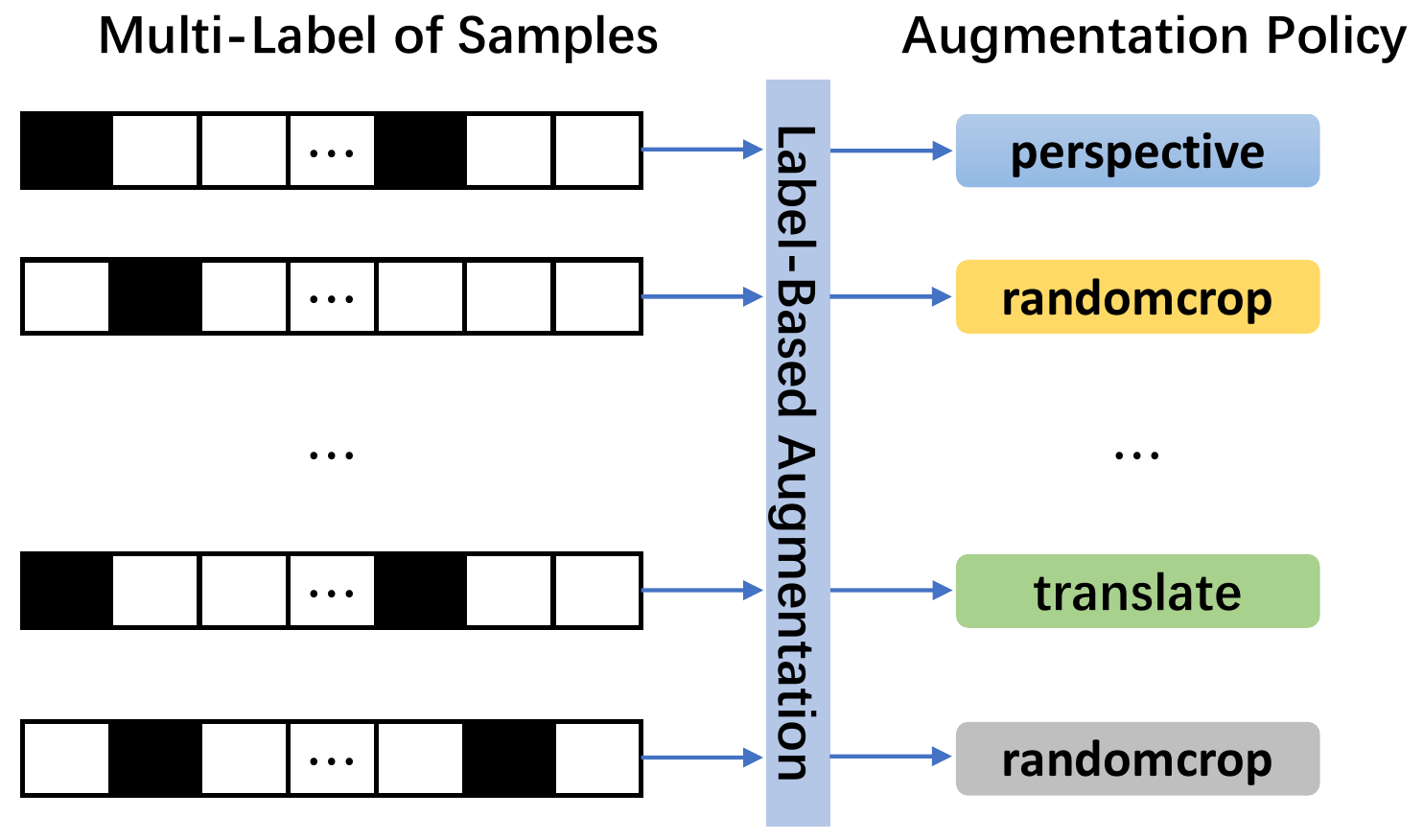} \label{fig:overall}}
	\caption{(a) The effects of some examplary augmentation operators on different labels of Peta, where ``P'' indicates positive effects on the classification, ``N'' represents negative effects and ``-'' means neutral or inconspicuous influences. (b) The idea of LB-Aug: learning policies mapping encoded instance labels to their augmentation policies (probability and settings of each operator).}
\end{figure}

%most of them are label-agnostic. That is, the agent samples one augmentation and then applies it to an training instance, no matter what class that training instance belongs to. This problem is more severe in multi-label classification, where the data of different classes are imbalance, resulting in a sub-optimal set of transformations that could be applied to long-tailed instances.

% This problem is more severe in multi-label classification where different labels are sensitive to different augmentations, resulting in an empty set of transformations that could be safely applied to all instances.

% Despite the great success of the existing learning-based augmentation policies, most of them are label-agnostic. 
However, most of the aforementioned methods are label-agnostic, where augmentation policies are determined regardless of what class a training instance belongs to.
This kind of policies are observed problematic in multi-label classification tasks in our study on the effects of augmentation policies on different labels, as shown in Figure~\ref{fig:influence}. The study is conducted using 7 augmentation operators (the row coordinates of Figure~\ref{fig:influence}) on Peta, a pedestrian attribute recognition dataset~\cite{deng2014pedestrian}.
% To get a concrete example, we conduct an experiment on Peta, a pedestrian attribute recognition dataset. 
% In the experiment, 7 augment operations is chosen for illustration.
The values in the table indicate the effects of each single operator on different labels (the column coordinates), with ``P'', ``N'' and ``-'' represent positive, negative and neutral effects respectively.
A positive effect means a model trained with a certain augmentation operator obtains a better mean accuracy (larger than a threshold) in a certain label than that of a model trained without.
% We first apply no operations to instances to train a classifier. Then for each operation, we independently add it from the pre-processing pipeline and then train a new classifier. 
% The performance differences between the two classifiers tell us whether a specific augment operation is useful or not, or even harmful.
% Results are shown in Figure~\ref{fig:influence}. 
% For each task (class) and operation, if the performance gain of using this operation in predicting this class is more than a threshold, we mark it by $1$; if opposite, we mark it by $-1$; if no significant gain or loss, we mark it by $0$. 
From Figure~\ref{fig:influence}, it is interesting to see that ``perspective'' is beneficial to the prediction of global ``Casual Upper'' and ``Jacket'', but harmful to ``Long Hair'' and ``No Carrying''. Meanwhile, the accuracy of ``Formal Upper'' is increased by ``randomcrop'' but decreased by ``scale''.
These results indicate that a certain augmentation operator may have opposite effects on different labels. In other words, label-agnostic policies are not necessarily beneficial to all labels.
% These indicate that certain augment operator does not always contribute to any task. 

%most of them are label-agnostic. That is, the agent samples one augmentation and then applies it to an training instance, no matter what class that training instance belongs to. This problem is more severe in multi-label classification, where the data of different classes are imbalance, resulting in a sub-optimal set of transformations that could be applied to long-tailed instances.

% In this work, we try to solve the above problem by exploiting instance label in the augmentation policy. 

% Therefore, when a label-agnostic method is used, the learned policy must be eclectic to all instances and tasks, resulting in sub-optimal performance. 
To address the above problem, we think instance labels must be considered in a fine-grained manner for data augmentation. To this end, a novel Label-Based AutoAugmentation (LB-Aug) method is proposed in this paper to learn label-aware policies (rather than label-agnostic ones), mapping from instance labels to their optimal data augmentation policies, as shown in Figure~\ref{fig:overall}.
% we follow the autoaugmentation framework to learn the optimal policy given the instance label. 
% We call this approach as fine-grained multi-label auto-augmentation.
The policies are learned through reinforcement learning using the Policy Gradient method~\cite{sutton2000policy}, where a policy is defined by the probabilities of applying each augmentation operator and their parameters.
The density matching scheme in FAA~\cite{lim2019fast} is also used here to speed up the policy learning process.
% The Policy Gradient is accelerated with density matching . 
The superiority of the method is demonstrated in extensive experiments on Peta, MS-COCO and Charades benchmarks. To the best of our knowledge, LB-Aug is the first label-based fine-grained autoaugmentation method
for multi-label classification.
% in computer vision. 

%In this paper, we aim to search for label based augmentation policies on the task of multi-label image classification and multi-label video recognition. In our implementation, taking 
The remainder of this work is organized as follows. First, relevant autoaugmentation methods are summarized in Section~\ref{relatedwork}. Then, the LB-Aug method is introduced in Section~\ref{methods}, followed with extensive experiments in Section~\ref{experiment} demonstrating the superiority of the method and showing how each part of the method contributes to the performance gain via ablation study.
% present our problem setting in formal and propose our Label Based AutoAugmentation . After that, the superiority of our method is demonstrated through extensive experiments in Section .
% Finally, we make detailed ablation analysis in Section \ref{ablation} to validate the effectiveness of LB-Aug.

\section{Related Work} \label{relatedwork}
Data augmentation plays an important role in improving the generalization of deep neural networks, especially for the task on small scale dataset. For example, random crop, flip, rotation, scaling, and color transformation have been performed as baseline augmentation methods in various tasks on ImageNet \cite{hu2018squeeze}, Kinetics \cite{feichtenhofer2019slowfast} and Charades \cite{wang2020multi}. Recently, plenty of augmentation methods have been proposed, such as Cutout \cite{devries2017improved} and Mixed-Example\cite{summers2019improved}. Though the augmentation policy is a key technique to enhance the generalization of deep neural networks, it is shown that different augmentation operations not always benefit their performance in different tasks. Besides, the hyper-parameters, such as mean and std in GaussianBlur, angle in the RandomRotate should also be chosen carefully. 
Different from common human designed augmentation operations, the methods of automatical augmentation take various augmentation operations to construct a search space and some sub-policies, and then automatically find the best augmentation policies through various learning strategies. AutoAugment \cite{cubuk2019autoaugment} introduces a reinforcement learning based search strategy that alternately trained a child model and RNN controller and leverages the performance of baseline DNN obviously. Although it is accelerated by PPO \cite{schulman2017proximal} method, the search process requires thousands of GPU hours. To reduce the searching cost, some improvements have been proposed.
%Taking the accuracies of classifier as rewards and accelerated by PPO \cite{schulman2017proximal} method, Autoaugment leverages the performance of baseline DNN obviously. 

Fast AutoAugmentation (FAA) takes Bayesian optimization as optimizer and the density matching as rewards to avoid training the classifier frequently, its training time is reduced remarkably in comparison with AutoAugment \cite{cubuk2019autoaugment}, which makes it an available method to be complemented in the task on large scale dataset. Adversarial AutoAugment \cite{zhang2019adversarial} is also an computationally-affordable solution, which attempts to increase the training loss of a target network through generating adversarial augmentation policies, while the target network can learn more robust features from harder examples to improve the generalization. PBA \cite{ho2019population} generates nonstationary augmentation policy schedules instead of a fixed one to accelerate the search for augmentation policies. In summary, previous learned augmentation methods mainly focus on reducing the searching computational complexity rather than improving the  performance and exploring new applications. Since they take augmentation operations and magnitudes into the evaluation without other information, all training samples share the same policies during training. It may be not optimal. On the other hands, the best augmentation policies are evaluated in only single-label classification tasks, e.g. CIFAR-10, CIFAR-100 and ImageNet. When transforming them into a multi-label classification tasks, we cannot obtain an expected result, which is shown in the previous experiment. In the next section, we will explain how do we solve these two major problems.

%However, as we know, our work is the first to deal with the problem that the influence of augmentation policies to different categories is discriminatory. In this paper, we propose a label based autoaugmentation methods to find a more fine-grained augmentation policy which is label specified.
\section{Method} \label{methods}
\begin{figure}[!t]
	\centering
	\subfigure[Training]
  {\includegraphics[width=0.53\textwidth]{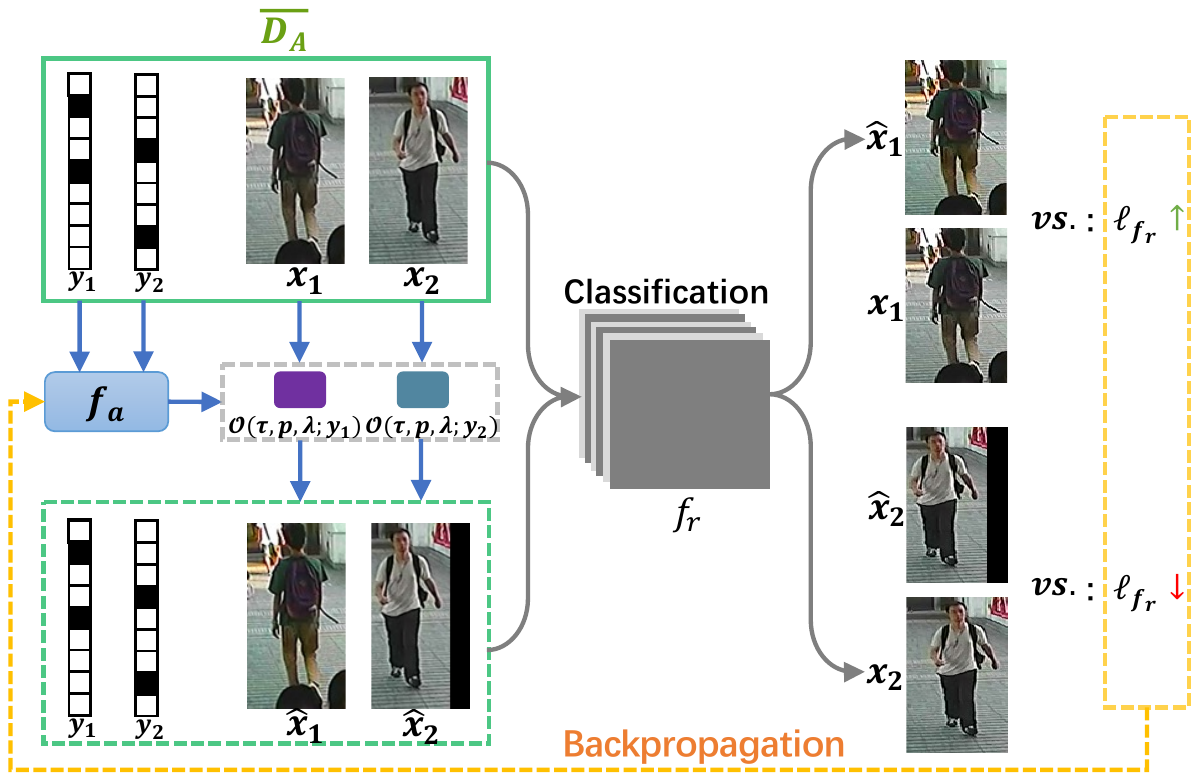} \label{train}} \quad
	\subfigure[Inference] 
  {\includegraphics[width=0.42\textwidth]{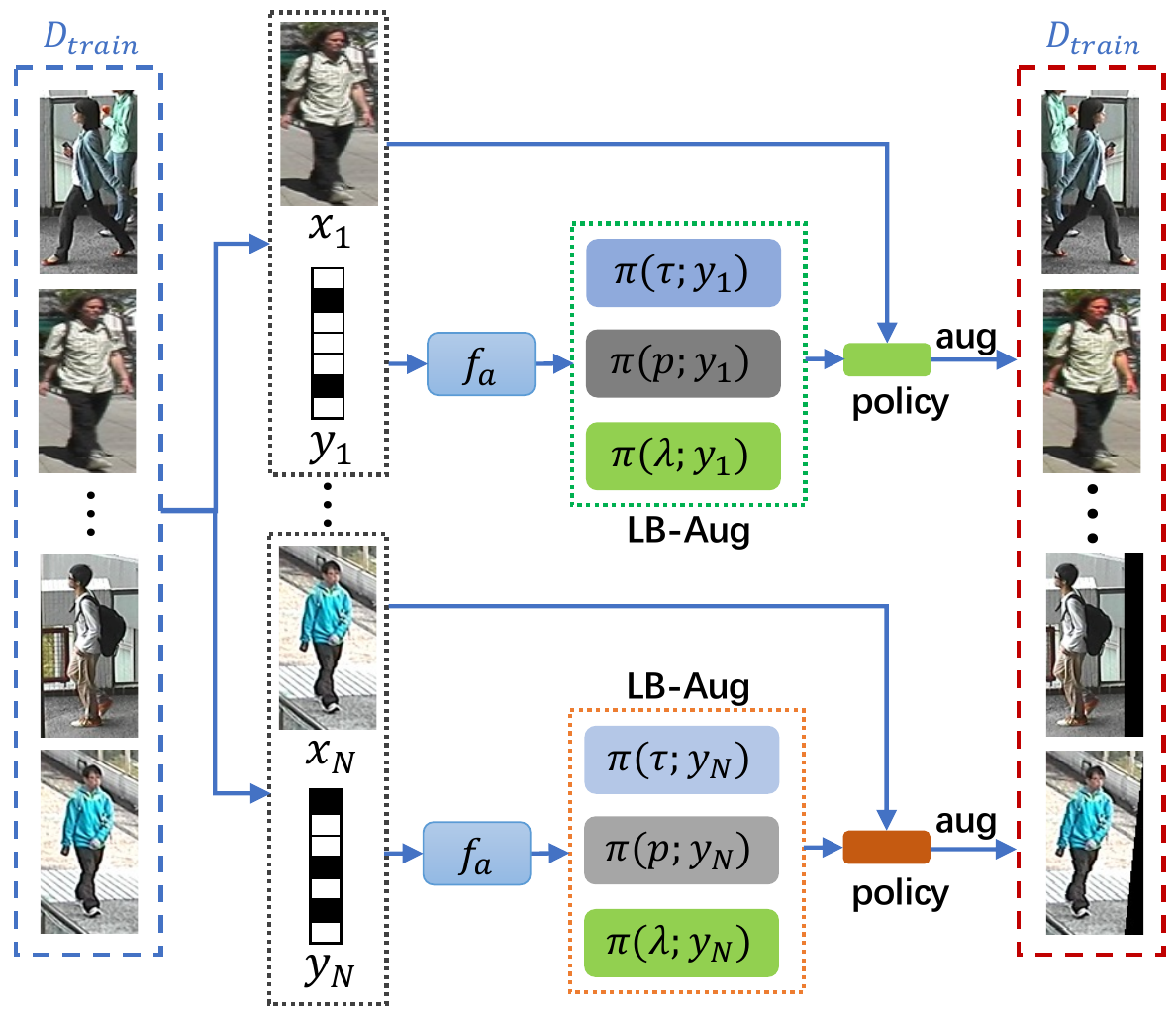} \label{validate}}
	\caption{(a) The training of the actor network $f_a$. For each instance in $\overline{D_A}$, its corresponding critic network $f_r$ evaluates the gain/drop of applying a testing augmentation policy on it. Then the comparison results are rewarded to $f_a$. In this way, $f_a$ updates its parameters through backpropagation. (b) An example of augmenting Peta training dataset via LB-Aug. The actor network takes the multi-hot label $y$ of each instance $x$ as input, then predicts specific probability distributions of operation $\tau$, calling probability $p$ and magnitude $\lambda$, each of which is label-based. After that, the label-based policy is conducted to augment the input instance $x$.}
\end{figure}

Our method provides fine-grained and label-based augmentation policies from a specific search space (Section \ref{searchspace}) for each image/video. The generic architecture involves two processes: 1) training an actor network with improved density matching (Section \ref{density}) as illustrated in Figure \ref{train}. 2) Using the trained actor network to construct the integral label based augmentation policy for each individual instance, shown in Figure \ref{validate}. 

\subsection{Label Based Search Space} \label{searchspace}
The search space in this paper contains 16 different operations per \cite{cubuk2019autoaugment}. Each operation $\tau$ requires a probability $p$ of being called, and a magnitude $\lambda$ with which it is applied. Formally, a label based augmentation policy $\mathcal{O} = (\tau, p, \lambda, y)$ for instance $(x, y)$ is :
\begin{equation} \label{apply}
	\mathcal{O}(x; p, \lambda, y) = \begin{cases}
		\tau(x; \lambda, y) &\text{with probability}\  p \\
		x& \text{with probability}\  1-p
	\end{cases}
\end{equation}
Note that, the calling probability $p$ and magnitude $\lambda$ in Eqn. \ref{apply} are both sensitive to the label $y$, where the ``Label Based'' derives from. We also discretize the ranges of calling probability into 11 values and magnitude into 10 values with uniform spacing, so that a discrete optimization algorithm (elaborated in Section \ref{density}) is available to facilitate the policy search phase.

% \subsection{Density Matching} \label{density}
% Density matching is an efficient mechanism to provide the reward of each policy evaluation without any backpropagation for network training. Given a pair of training and validation datasets $D_{train}$ and $D_{valid}$, the purpose is to facilitate the generalization ability of policies to match the density of $D_{train}$ and augmented $D_{valid}$. However, the density of these two distributions is not computationally available. Therefore, density matching performs this evaluation by measuring the generalization of the classifier pre-trained on one dataset to the other datasets. In detail, it first splits $D_{train}$ into two segments $D_M$ and $D_A$ that are used for learning a multi-label classifier $f_r$ (\emph{critic network}) and exploring the augmentation policy $\mathcal{T}$, respectively. The following loss is employed to find the optimal augmentation policies, 
% \begin{equation}
% 	\mathcal{T}_{*} = \underset{\mathcal{T}}{\text{argmax}}\ \mathcal{R}(f_r | \mathcal{T}(D_A))
% \end{equation}
% where $f_r$ is pre-trained on $D_M$. $\mathcal{T}(D_A)$ is the augmented datasets $\mathcal{T}(D_A) = \{\mathcal{T}(x) | x \in D_A\}$. $T_{*}$ minimizes the density gap of $D_M$ and $D_A$ from the perspective of enhancing the generalization ability of $f_r$ to $D_A$.

\subsection{Label-based Augmentation} \label{density}
From the perspective presented in Figure~\ref{fig:influence}, we propose learning a Label Based AutoAugmentation (LB-Aug) to yield fine-grained augmentation for multi-label classification. The term ``fine-grained'' implies that the policies are adaptive to different samples with their labels. 

\subsubsection{Training Actor with Improved Density Matching} \label{sec:improved}
Density matching~\cite{lim2019fast} is an efficient mechanism to provide the reward of each policy evaluation without any backpropagation for network training. Given a pair of training and validation datasets $D_{train}$ and $D_{valid}$, for the purpose of facilitating the generalization ability of policies to match the density of $D_{train}$ and augmented $D_{valid}$, density matching performs this evaluation by measuring the generalization of the classifier pre-trained on one dataset to the other dataset. In detail, it splits $D_{train}$ into two segments $D_M$ and $D_A$ those are used for learning a multi-label classifier $f_r$ (\emph{critic network}) and exploring the augmentation policy $\mathcal{T}$, respectively. The following loss function is employed to find the optimal augmentation policies, 
\begin{equation}
	\mathcal{T}_{*} = \underset{\mathcal{T}}{\text{argmax}}\ \mathcal{R}(f_r | \mathcal{T}(D_A))
\end{equation}
where $f_r$ is pre-trained on $D_M$. $\mathcal{T}(D_A)$ is the augmented datasets $\mathcal{T}(D_A) = \{\mathcal{T}(x) | x \in D_A\}$. $\mathcal{T}_{*}$ minimizes the density gap of $D_M$ and $D_A$ from the perspective of enhancing the generalization ability of $f_r$ to $D_A$.

Normally, the scales of $\mathcal{T}(D_A)$ is several times smaller than the original $D_{train}$, limiting the generalization the learnt $\mathcal{T}_{*}$. To this end, we propose to improve the density matching by enlarging the training set of $\mathcal{T}$ with full use of $D_{train}$. Considering the original training set $D_{train}$, the improved density matching uses K-fold stratified shuffling \cite{shahrokh2014effect} to get $K$ segmentations $\{S^{(1)}, \cdots, S^{(K)}\}$, each $S^{(k)}$ consists of two datasets $D_M^{(k)}$ and $D_A^{(k)}$, where $D_{train}=D_M^{(k)} \bigcup D_A^{(k)}$. Following that, we first train a multi-label classifier $f_r^{(k)}$ on each $D_M^{(k)}$. Then, we get an unified evaluation set 
\begin{equation} \label{unions}
\overline{D_A}=\bigcup_{k=1}^{K} D_A^{(k)}
\end{equation}
Assuming an instance $x\in D_A^{(k)}$, the multi-label classifier $f_r^{(k)}$ trained on $D_M^{(k)}$ is thought as its corresponding critic network. Obviously, each instance in $\overline{D_A}$ has at least one corresponding critics.  We introduce the training of actor network taking $\overline{D_A}$ as the training set, which is described next.

Given an instance $(x_i, y_i)$ in $\overline{D_A}$, the searching of its optimal policy is based on Policy Gradient \cite{sutton2000policy}, which has two terms: 1) an actor network $f_a$ predicts the optimal policy $\mathcal{O} = (\tau, p, \lambda, y_i)$ while taking label $y_i$ as input. 2) the corresponding critic networks $\{f_r^{(k)}|x_i \in D_A^{(k)}, \forall k\}$ to reward $\mathcal{O}$, as illustrated in Figure \ref{train}. In detail, actor network $f_a$ provides an triple probability distributions
\begin{equation} \label{actorwork}
	(\pi(\tau;y_i), \pi(p;y_i), \pi(\lambda;y_i)) = f_a(y_i)
\end{equation}
Here $\pi(\tau;y_i) \in \mathbb{R}^{C_{\tau}}$is a vector that indicates the probability distribution of calling each operation. Each row of $ \pi(p;y_i) \in \mathbb{R}^{C_{\pi}\times C_p}$ and $\pi(\lambda;y_i) \in \mathbb{R}^{C_{\pi}\times C_{\lambda}}$ presents the probability distributions of aforementioned discretized calling probabilities and magnitudes, respectively. During inference, a categorical distribution is conducted with $\pi(\tau;y_i)$ to sample an augmentation operation, for example $\tau_j$. Its calling probability and magnitude are then sampled with probabilities $\pi_j(p ; y_i)$ and $\pi_j(\lambda ; y_i)$.

To maximize the gain of sub-policies, a form of difference loss is considered to train the actor network $f_a$: 
\begin{equation} \label{loss}
	\mathcal{L} = \frac{1}{N}\sum_{i=1}^{N} \mathbb{E}_{jkl}\,  log\, \pi(\tau_j, p_k,\lambda_l; y_i) \Delta \ell
\end{equation}
For simplification, the calling probability and magnitude are considered independent from each other, so that the chain rule is available:
\begin{equation}
	\pi(\tau_j, p_k,\lambda_l ; y_i) =\pi(\tau_j; y_i)\pi_j(p_k ; y_i) \pi_j(\lambda_l ;y_i)
\end{equation}
$\Delta \ell$ is the reward considered as the gain of classification loss from multi-label critic network $f_r^{(k)}$
\begin{equation}
	\Delta \ell = \mathbb{E}_{x_i \in D_A^{(k)}} \left[ \ell_{f_r^{(k)}}(x_i, y_i) - \ell_{f_r^{(k)}}(\hat{x}_i, y_i) \right] \
\end{equation}
$\ell_{f_r^{(k)}}$ is a binary cross entropy loss between the prediction of $x_i$ and ground-truth label $y_i$. $\hat{x}_i$ is the augmented counterpart of $x_i$ and $\hat{x}_i=\tau_j(x_i; p_k, \lambda_l)$. Intuitively, the loss $\mathcal{L}$ increases the probability of group $(\tau, p, \lambda)$ which leads to better performance gain, and vice versa.

% \begin{figure}[!t]
% 	\centering
% 	\includegraphics[width=\columnwidth]{figures/train.pdf}
% 	\caption{The overview of augmentation search by LB-Aug. The training set $D_{train}$ is splitted into a $D_M$ where an multi-label classifier $f_r$ is pre-trained and a $D_A$ to train an actor network $f_a$. Taking images after and before the augmentation being applied, the classifier $f_r$ acts as a critic and evaluates whether the classification loss $\ell_{f_r}$ is increased or decreased. Then the reward is collected to the backpropagation of learning $f_a$.}
% 	\label{fig:framework}
% \end{figure}

\subsubsection{Label-based Augmentation for Multi-label Classification}
Figure \ref{validate} presents the usage of actor networks learned preliminarily. Taking the label $y_i$ as input, a label-based augmentation policy is predicted for each instance $(x_i, y_i) \in D_{train}$ by the well-trained actor network $f_a$. It could remedy the distribution gap of original training set $D_{train}$ and validation set $D_{valid}$ more flexibly than label-agnostic ones. In this way, the proposed LB-Aug enhances the generalization of the final multi-label classifier $f_c$ (having the same architecture as $f_r$ by default).  

A unique advantage of LB-Aug is that it is embedded with label-based information so that the augmentation of instances is not fixed globally but adaptive to each individual. In this way, it is superior to handle the severe problem faced by multi-label tasks that different labels are sensitive to different augmentations.

\section{Experiments} \label{experiment}

Extensive experiments are first conducted to evaluate the effectiveness and superiority of the LB-Aug by comparing it with some baseline methods (Section~\ref{sec:baselines}) and previous SOTA approaches (Section~\ref{sec:sota}), followed with a number of ablation studies showing how the method contributes to the performance gain of each label (Section~\ref{sec:label_wise_analysis}) and how its performance is influenced by different settings on RL frameworks and the number of folds for density matching (Section ~\ref{reinfor} to~\ref{sec:number_of_folds}).

% ~\ref{sec:net_depth}~\ref{sec:rl_frameworks},~\ref{sec:number_of_folds}

% \subsection{Experimental Results}
% The proposed LB-Aug can seamlessly be injected into the training of any common deep learning network and improve its performance for multi-label image/video recognition tasks. In this part, we first demonstrate that our method surpasses other augmentation based methods by large margins. Then the experimental comparison on Peta shows that standard Inception-V3 with LB-Aug could yields state-of-the-art results with obvious improvements over the previous superior methods. Finally, we conduct sufficient ablation studies.

% We first show that LB-Aug can improve the performance of baseline multi-label classifiers dramatically without any impact on the inference speed. 
% Then 
% Notably, It further demonstrated state-of-the-art performance on Peta. 
% Then, a number of ablation studies are  to evaluate our LB-Aug. In general we note that it is robust to the exact design of the Reinforcement Learning methods, while the variation of output policies shows profound difference in our experiments. 

\subsection{Experimental Setup}
\noindent\textbf{Dataset and Evaluation Metrics}
We evaluate our method on three image/video datasets with standard evaluation protocols. For multi-label image classification task, we use Peta \cite{deng2014pedestrian} and MS-COCO \cite{Lin2014Microsoft} datasets, report the standard attribute-wise mean accuracy (mA), overall accuracy (Accu), F1 for Peta and mAP, per-class F1~(CF1), overall F1~(OF1) for MS-COCO. For the multi-label video recognition experiment, we consider the widely used Charades \cite{Sigurdsson2016Hollywood} and only mAP is evaluated, following \cite{wang2020multi}.

\emph{Peta} is a large-scale pedestrian attributes dataset containing 19.0K images among which 9.5K is used for training, 1.9K for validation and 7.6K for model evaluation \cite{tang2019improving}. It has 61 binary attributes and 4 multi-class attributes. While, we only fetch 35 attributes whose positive proportions are larger than 5\% for model analysis. \emph{MS-COCO} is also a widely used image classification baseline. It contains about 82.1K images for training, 40.5K for validation and 40.7K for test. On average, each image has 2.9 labels from a set of 80 object labels. \emph{Charades} has 157 action classes containing around 9.8K daily indoors activities videos, where 8.0K is for training and 1.8K for validation. Each video is labeled with 6.8 actions on average.

\noindent\textbf{Augmentation Operations}
The search space consists of 16 operations (ShearX, ShearY, TranslateX, TranslateY, Rotate, AutoContrast, Invert, Equalize, Solarize, Posterize, Contrast, Color, Brightness, Sharpness, Cutout and Sample Pairing). The magnitude of each operation follows the default ranges in AutoAugment (AA) \cite{cubuk2019autoaugment} and Fast AutoAugment (FAA) \cite{lim2019fast}.

\noindent\textbf{Architecture}
By default, our experiments use standard ResNet50, ResNet101 and Inception-V3 for image classification tasks, Inception-I3D and S3D for video recognition tasks. The outputs are activated by Sigmoid and learned by joint binary cross entropy loss. For actor networks, we concatenate three fully connected layers with Dropout rate 0.5 and ReLU superimposing.

\noindent\textbf{Training Details}
The training of LB-Aug is boosted by the improved density matching (Section \ref{sec:improved}). Unless noted otherwise, we split training datasets into 8 folds with Stratified Shuffling \cite{lim2019fast}, where $4/5$ of original training set is segmented as $D_M$ and the rest as $D_A$. For Peta and MS-COCO, we use SGD as optimizer with a mini-batch size of 80 and initial learning rate of 0.001. The weight decay is $10^{-4}$ and momentum is 0.9. For Charades, each crop is randomly cropped into $224 \times 224$ from a $256\times 256$ scaled video. Adam with initial learning rate 0.001 and weight decay $10^{-4}$ is used as optimizer for better convergence. The mini-batch size is 8 clips, while the accumulated gradient \cite{lin2017deep} of step 8 is utilized to stabilize the model training.

\subsection{Comparisons with Baselines}
\label{sec:baselines}
The quantitative evaluation on multi-label image classification is shown in Table \ref{peta2} and \ref{msscoco}, and multi-label video recognition in Table \ref{charades_result}. For each table, we compare the gain of Random policies (noted by ``Baseline+Random''), FAA, our LB-Aug$_E$ (with fixed hyper-parameters) and LB-Aug$_H$ (with trainable calling probabilities and magnitudes) over baseline models. In this paper we omit comparing our result with AA for two aspects: 1) AA \cite{cubuk2019autoaugment} is computationally consuming and the training of augmentation policies takes thousands of GPU hours. 2) FAA is an accelerated alternative to AA with little performance gap, which is demonstrated in \cite{lim2019fast}. Additionally, the released experiments on FAA \cite{lim2019fast} is limited in single label tasks. For fair comparison on multi-label tasks, we conduct FAA with the released code of \cite{lim2019fast} and follow its best setting.

The training and testing protocols are kept the same for all methods except that the training of FAA follows Bayesian optimization \cite{lim2019fast}. 

\begin{table}[!ht]
	% \centering
	% \footnotesize
	%\setlength{\tabcolsep}{1.5mm}
	\caption{Performance comparison on Peta among baselines and LB-Aug.}
	\label{peta2}
	\resizebox{\columnwidth}{!}{
	\begin{tabular}{c|c|c|c|c|c|c|c|c|c}
	\hline
	\multirow{2}{*}{Method} & \multicolumn{3}{c|}{ResNet50} & \multicolumn{3}{c|}{ResNet101} & \multicolumn{3}{c}{Inception-V3} \\
	\cline{2-10}
	& mA & Accu & F1 & mA & Accu & F1  & mA & Accu & F1\\
	\hline
	Baseline$^{\dagger}$ & 84.9 & 78.1& 85.5 &85.4&78.9&85.8 &86.0 &79.6 &86.5 \\
	Baseline+Random$^{\ddagger}$ & 85.2 &78.5&85.2&85.6&79.0&86.0 &86.2 &79.7 &86.5 \\
	FAA \cite{lim2019fast}  & 85.5 &78.9 &85.7&85.8&79.3&86.4 &86.4 &79.8 &86.8 \\
% 	LB-Aug$_E$$^{\ast}$ (DM$^{+}$)  & 86.5 &79.7&86.8&86.6 & 80.0&86.8 &87.1 &80.7 &87.3 \\
% 	LB-Aug$_H$$^{\star}$ (DM)  &\textbf{86.8} &80.0 &86.9 &86.9 &80.3 &\textbf{87.1} &87.3 &\textbf{80.9} &87.5 \\
	LB-Aug$_E$$^{\ast}$  & 86.6 &79.9&86.8&86.6 & 80.1&86.8 &87.2 &80.8 &87.3 \\
	LB-Aug$_H$$^{\star}$  &\textbf{86.8} &\textbf{80.1} &\textbf{87.0} &\textbf{87.0} &\textbf{80.4} &\textbf{87.1} &\textbf{87.4} &\textbf{80.9} &\textbf{87.6} \\
	\hline
	\multicolumn{10}{l}{$^{\dagger}:$ counterpart without additional augmentation from search space.} \\
	\multicolumn{10}{l}{$^{\ddagger}:$ counterpart with random policies from search space.} \\
	\multicolumn{10}{l}{$^{\ast}:$ counterpart with fixed calling probabilities 0.5 and magnitudes 1.} \\
    \multicolumn{10}{l}{$^{\star}:$ counterpart with learnable calling probabilities and magnitudes.}
	\end{tabular}}
\end{table}
  
\begin{table}[!ht]
	% \centering
	% \footnotesize
	%\setlength{\tabcolsep}{1.5mm}
	\caption{Comparisons between baselines and LB-Aug on MS-COCO.}
	\label{msscoco}
	\setlength\tabcolsep{2.2mm}
	\resizebox{\columnwidth}{!}{
	\begin{tabular}{c|c|c|c|c|c|c|c|c|c}
	\hline
	\multirow{2}{*}{Method} & \multicolumn{3}{c|}{ResNet50} & \multicolumn{3}{c|}{ResNet101} & \multicolumn{3}{c}{Inception-V3} \\
	\cline{2-10}
	& mAP & CF1 & OF1 & mAP & CF1 & OF1  & mAP & CF1 & OF1 \\
	\hline
	Baseline &74.7&69.4&73.8& 77.1 &71.3 &76.0 & 77.6& 74.1&76.4\\
	Baseline+Random &75.1&70.6&74.1& 77.4 &72.8 &76.7&78.2&74.4&76.7\\
	FAA &76.0&71.1& 75.3&  78.7&73.5 &76.4&79.1&74.7&76.9 \\
% 	LB-Aug$_E$ (DM) &76.4&71.2 &75.9& 79.1 &74.2 &77.3 &79.3 & 75.0 & 77.5\\
% 	LB-Aug$_H$ (DM) &\textbf{77.1}&\textbf{71.4} &\textbf{76.3}&\textbf{79.5} &\textbf{74.8} &\textbf{77.6} &\textbf{79.7}&\textbf{75.1}&\textbf{77.7}\\
	LB-Aug$_E$ &77.1&72.3 &76.2& 79.7 &74.5 &77.5 &80.2 & 75.9 & 77.8\\
	LB-Aug$_H$ &\textbf{77.4}&\textbf{72.6} &\textbf{76.5}&\textbf{80.1} &\textbf{74.8} &\textbf{77.8} &\textbf{80.4}&\textbf{76.2}&\textbf{78.0}\\
	\hline
	\end{tabular}}
\end{table}

\begin{table}[!ht]
\caption{Performance for multi-label video recognition on Charades.}
		\label{charades_result}
	\centering
	\setlength\tabcolsep{4mm}
	\resizebox{\columnwidth}{!}{
		\begin{tabular}{c|c|c|c|c|c|c}
		\hline
		\multirow{2}{*}{Backbone} & \multirow{2}{*}{Baseline} & \multicolumn{3}{c|}{Baseline+Random} & \multirow{2}{*}{FAA}& \multirow{2}{*}{LB-Aug}  \\
		\cline{3-5}
		& & $p^{\dagger}=0.25$ & $p=0.75$ & $p=1$ & \\
		\hline
		I3D & 36.3 & 35.5 & 34.8 & 34.1 &36.4 & 37.6\\
		S3D & 36.8 & 36.2 & 35.7 & 35.0 &37.0 & \textbf{38.0} \\
		\hline
		\multicolumn{7}{l}{$^{\dagger}:$ the magnitude of augmentation operations.}
		\end{tabular}}
\end{table}

\noindent\textbf{Multi-label Image Classification}
Table \ref{peta2} and Table \ref{msscoco} show the comparison results with various augmentation methods and backbones: ResNet50, ResNet101 and Inception-V3. In this case, randomly chosen augmentation policies could lead to better generalization of all three backbones due to a suitable search space for these tasks, nevertheless the gains is relatively trivial. Compared with previous state-of-the-art FAA, our two counterparts LB-Aug$_E$ and LB-Aug$_H$ both provide substantially improvement over all evaluation matrics. In particular, FAA and LB-Aug$_H$ both adaptively learn the hyper-parameters, while with vital label information embedded, our LB-Aug$_H$ leads to over 1.0\% mA, 1.1\% Accu and 0.7\% F1 in Peta, as well as over 1.3\% mA, 1.3\% CF1 and 1.1\% OF1 in MS-COCO.

\noindent\textbf{Multi-label Video Recognition}
For the task of multi-label video recognition, we show the results on two backbone models: I3D and S3D in Table \ref{charades_result}. The search space is inherited from Peta and MS-COCO, except that the videos are augmented frame by frame. Contrary to the results on Peta and MS-COCO, random augmentation policies drop over 0.8\% mAP due to the amateurish search space for video tasks. While both FAA and our LB-Aug could still get gains over baseline. We notice that benefited from a fine-grained searching mechanism, our LB-Aug outperforms FAA by solid margins (37.6\% \textit{vs.} 36.4\% for I3D, 38.0\% \textit{vs.} 37.0\% for S3D).

\subsection{Comparisons with State-of-the-arts}
\label{sec:sota}
The proposed LB-Aug achieves two new SOTA performances on both Peta and MS-COCO with more strong backbone models. Table \ref{peta} presents the comparison results of LB-Aug against other SOTA methods on Peta. In comparison to the previous SOTA work GRL~\cite{zhao2018grouping}, our best model suggests up to 0.7\% higher mA and 1.1\% higher F1 while with the same backbone Inception-V3. In additional, GRL makes gains over backbone Inception-V3 by utilizing extra body region proposal and a Recurrent Neural Network (RNN), which enlarges the computation cost. Instead, the result of our proposed LB-Aug is achieved only taking backbone Inception-V3 as classifiers without other tricks on model design. On the experiments on MS-COCO, with the SOTA backbone model TResNet-L, our model achieves a 0.3\% gain on mAP. Under the same experimental setting, the gain is up to 0.6\%. Specifically, the gains on Peta and MS-COCO are both made by leveraging the label information for data augmentation without any extra modification on algorithm itself. We emphasize that our mechanism could be a more naive way to get bonus from other state-of-the-art works.

\begin{table}[!ht]
  \centering
  \caption{Comparing LB-Aug against other SOTA methods on Peta.}
    \label{peta}
  \setlength\tabcolsep{5mm}
  \resizebox{\columnwidth}{!}{
    \begin{tabular}{c|c|c|c|c|c}
      \hline
      Methods & Backbone & Pretrain & mA & Accu &F1 \\
      \hline
	  JRL \cite{wang2017attribute} &AlexNet &ImageNet &85.7& $-$& 85.4\\
% 	  ACN \cite{sudowe2015person}    &CaffeNet  & ImageNet      & 81.2& 73.7& 82.6 \\
	  DeepMar \cite{li2015multi} &CaffeNet  &ImageNet &82.9& 75.1& 83.4 \\
	  DeepMar$\dagger$ \cite{li2015multi} &Inception-V3  &ImageNet &81.5&$-$ & 85.7 \\
	  VeSPA \cite{sarfraz2017deep} &GoogleNet &ImageNet &83.5& 77.7&85.5 \\
% 	  HP-Net \cite{liu2017hydraplus} &Inception-V2 &$-$ &81.8& 76.1&  84.1 \\ 
	  WPAL \cite{zhou2017weakly} &GoogleNet & ImageNet & 85.5 & 77.0 & 84.9 \\
	  PGDM \cite{dangwei2018pose} &CaffeNet & ImageNet& 83.0& 78.1&85.8 \\
	  ALM \cite{tang2019improving} &BN-Inception & $-$&86.3& 79.5&  86.9 \\
	  GRL \cite{zhao2018grouping} &Inception-V3 &ImageNet & 86.7& $-$& 86.5 \\
	  FAA \cite{lim2019fast}  & Inception-V3 &ImageNet&86.4 &79.8 &86.8 \\
	  \hline
	  LB-Aug$_H$ & BN-Inception & ImageNet &86.7 &80.1 &87.2 \\
	  LB-Aug$_H$ & Inception-V3 & ImageNet &\textbf{87.4} &\textbf{80.9} &\textbf{87.6} \\
	  \hline
	  \multicolumn{6}{l}{$^{\dagger}:$ reported by \cite{zhao2018grouping}.}
    \end{tabular}}
  \end{table}
  
\begin{table}[!ht]
	% \centering
	% \footnotesize
	%\setlength{\tabcolsep}{1.5mm}
	\caption{Experimental comparison of SOTA methods on MS-COCO.}
	\label{coco}
	\setlength\tabcolsep{1.5mm}
	\resizebox{\columnwidth}{!}{
	\begin{tabular}{c|c|c|c|c|c}
	\hline
	Methods & Backbone & Pretrain &  mAP & CF1 & OF1 \\
	\hline
	SRN~\cite{zhu2017learning} &ResNet101 &ImageNet &77.1 & 71.2 & 75.8 \\
	Multi-Evidence~\cite{ge2018multi} &ResNet101 & ImageNet& $-$ & 74.9  & 78.4 \\
    CADM  \cite{chen2019multi} &ResNet101 &ImageNet &82.3&77.0&79.6\\
    ML-GCN \cite{chen2019multi2}&ResNet101 & ImageNet& 83.0&78.0&80.3\\
    KSSNet \cite{wang2020multi}&ResNet101 & ImageNet&83.7&77.2&81.5\\
    MS-CMA \cite{you2020cross} & ResNet101 & $-$ &83.8&78.4&81.0\\
    % MCAR& \cite{gao2020multi} 83.8&78.0&80.3\\
    ASL \cite{ben2020asymmetric}& TResNet-L&ImageNet &86.6\ (86.4$^{\star}$)&81.4\ (81.1$^{\star}$)&81.8\ (81.6$^{\star}$)\\
    ASL \cite{ben2020asymmetric}$^{\dagger}$& TResNet-L&ImageNet &88.4\ (88.1$^{\star}$)&$-$\ (81.6$^{\star}$)&$-$\ (82.3$^{\star}$)\\
    \hline
    LB-Aug$_H^{\ddagger}$ & TResNet-L&ImageNet&86.9& 81.6 &81.9\\
	LB-Aug$_H^{\dagger\ddagger}$ & TResNet-L&ImageNet&\textbf{88.7}&\textbf{82.3}&\textbf{82.7}\\
	\hline
    \multicolumn{6}{l}{$^{\star}:$ our implemented results with the released code of \cite{ben2020asymmetric}.}\\
	\multicolumn{6}{l}{$^{\dagger}:$ input resolution is enlarged to $640\times640$}.\\
	\multicolumn{6}{l}{$^{\ddagger}:$ conducted following the released code of ASL \cite{ben2020asymmetric}}.
	\end{tabular}}
\end{table}

\subsection{Ablation Studies} \label{ablation}

\subsubsection{Label-wise Performance}
\label{sec:label_wise_analysis}
\begin{figure}[!t]
	\centering
	\includegraphics[width=\columnwidth]{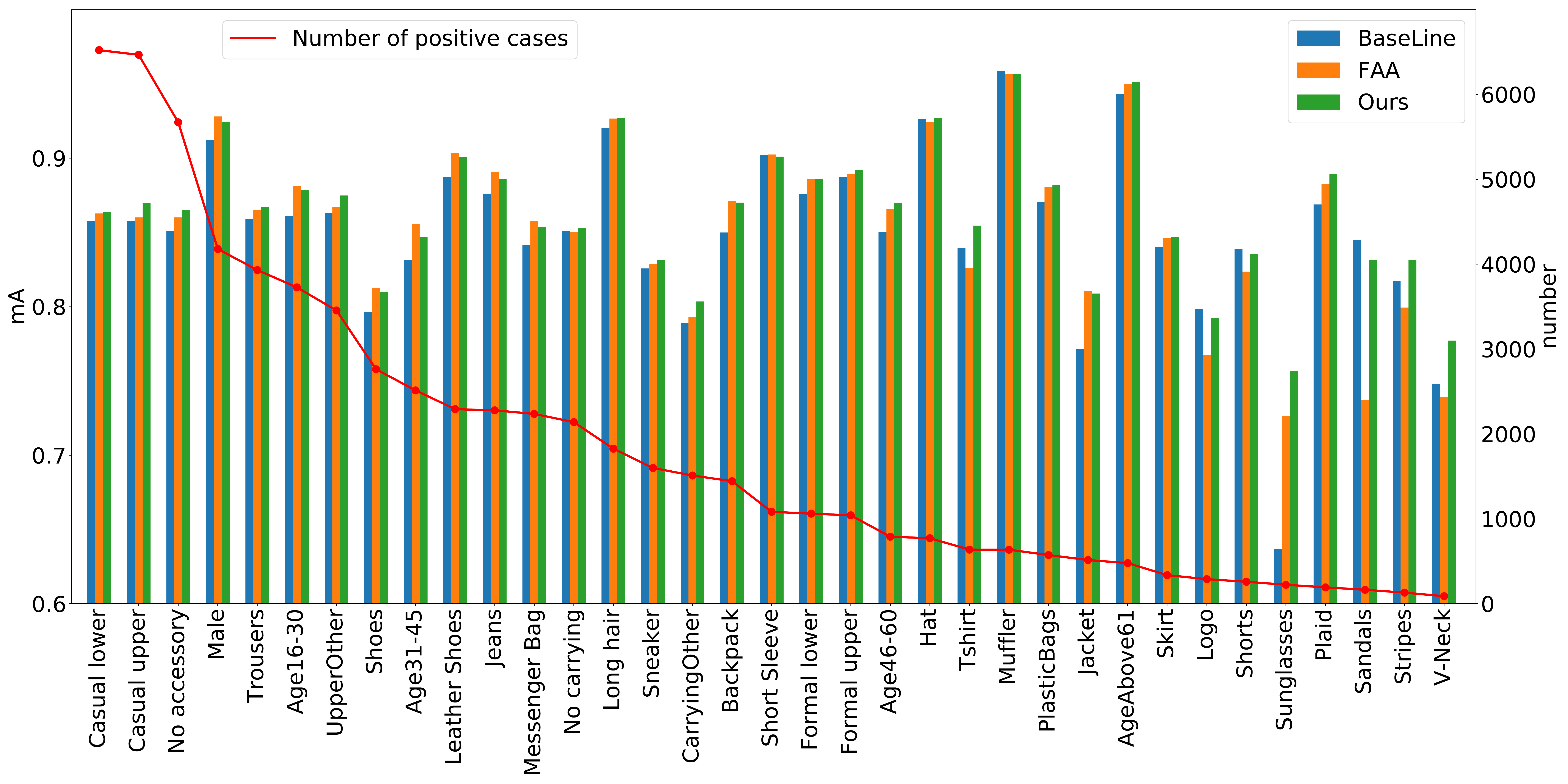}
	\caption{The mA histogram and positive case number of each label on Baseline ResNet50, FAA and LB-Aug.}
	\label{changwei}
\end{figure}

Figure~\ref{changwei} shows the performance bonus of FAA and LB-Aug per label. FAA gets an mA bonus of 0.6\% over baseline while benefited from learning better augmentation operations and hyper-parameters. However, compared to Baseline, its gain of overall mA mainly sources from enhancing the model performance on labels with more positive cases. As demonstrated in Table \ref{changwei}, it even drops its mA on labels with no sufficient positive rate. The intrinsic reason is that FAA is label-agnostic, meaning that it has to provide a compromised policy for better overall mA. Leveraged by more fine-grained and label-based search mechanism, our LB-Aug gets competitive performance against FAA on labels with large positive cases, while still keep performance gain over labels with small positive rate. 

% \todo[inline]{Add some experiments showing how each component contribute to the performance improvement, e.g., FAA vs. FAA + LB-Aug with 8 models vs. FAA + LB-Aug on one model. These results can be shown in a new table or by adding one more case in Figure 4.}

\subsubsection{Generalization to Other Models}
\label{sec:generalization}
We also show that a well trained policies from specific critic models can generalize to other models. Table \ref{peta3} compares the evaluation results of ResNet101 and Inception-V3 with a fixed policy which is learned from taking critic network as ResNet50. It is shown that a well trained policy consistently improves the performance of other models by negligible margin with retrained ones. 

\begin{table}[!ht]
	\caption{Comparing the mAP of fixed and retrained policy counterparts. The fixed policy is learned with critic model of ResNet50. The value in parentheses denotes the gain/drop over Table \ref{peta2}.}
	\label{peta3}
	\resizebox{\columnwidth}{!}{
	\begin{tabular}{c|c|c|c|c|c|c}
	\hline
	\multirow{2}{*}{Method} & \multicolumn{3}{c|}{ResNet101} & \multicolumn{3}{c}{Inception-V3} \\
	\cline{2-7}
	& mA & Accu & F1  & mA & Accu & F1\\
	\hline
	LB-Aug$_E$ &86.6 \textcolor{green}{(+0)}& 80.0 \textcolor{red}{(-0.1)}&86.6 \textcolor{red}{(-0.2)} &87.4 \textcolor{green}{(+0.2)} &80.9 \textcolor{green}{(+0.1)}&87.3 \textcolor{green}{(+0)}\\
	LB-Aug$_H$ &86.9  \textcolor{red}{(-0.1)} &80.3 \textcolor{red}{(-0.1)} &87.0 \textcolor{red}{(-0.1)} &87.5 \textcolor{green}{(+0.1)} &80.9 \textcolor{green}{(+0)} &87.5 \textcolor{red}{(-0.1)} \\
	\hline
	\end{tabular}}
\end{table}

% \subsubsection{Depth of Actor Network}
% \label{sec:net_depth}
% The actor networks of our LB-Aug is constructed with fully connected layers which are equipped with ReLU to make an non-linear transformation and Dropout to overcome the overfitting problem. Figure \ref{layers} illustrates the results of layer depths from 2 to 5. Depth of three is significantly better than depth 2, while as depth goes deeper, the performance on MS-COCO drops slightly, while drops more distinctly on Peta. The reason we think is the task of finding augmentation policy for each label is much more simple compared with common deep learning tasks. As the input of actor networks is a multi-hot vector whose length equals to the number of labels and the length of output is determined by the number of different operations (16 in this paper), the task of searching optimal policies for different labels is actually a simple classification, and suffers from ``overfitting'' problem. 

% \begin{figure}[!t]
%   \centering
%   \subfigure[mA on Peta] 
%   {\includegraphics[width=0.48\columnwidth]{./figures/layer_ma.png}}\ 
%   \subfigure[mAP on MS-COCO]
%   {\includegraphics[width=0.48\columnwidth]{./figures/layer_map.png}}
% 	\caption{mA on Peta and mAP on MS-COCO. Each fully connected layer is followed by an ReLU and Dropout whose drop rate is 0.5.}
% 	\label{layers}
% \end{figure}

\begin{table}[!ht]
		\caption{Quantitative results on Peta of different reinforcement learning framework. All counterparts are based on ResNet50.}
		\label{reinforcement}
	\centering
	\setlength\tabcolsep{9mm}
		\begin{tabular}{c|c|c|c}
		\hline
		Framework & mA & Accu & F1 \\
		\hline
		Policy Gradient \cite{sutton2000policy} & \textbf{86.8} &\textbf{80.1}&\textbf{87.0}\\
		DPG \cite{mnih2013playing} & 86.7 &80.0&\textbf{87.0} \\
		AC \cite{konda2000actor} & \textbf{86.8} &80.0&86.9 \\
		DQN \cite{silver2014deterministic} & 86.2 &79.6&86.7 \\
		\hline
		\end{tabular}
\end{table}

\subsubsection{Reinforcement Learning Framework} \label{reinfor}
\label{sec:rl_frameworks}

As we know, reinforcement learning suffers from the problem of convergence and sub-optimization, therefore different optimizers seems to have influence on the performance of LB-Aug. In order to analyze this influence, we conduct ablation study on the training framework of actor network on Peta dataset. Policy Gradient is used as the final reinforcement learning methods. For comparison, we also implements LB-Aug trained with AC \cite{konda2000actor}, DPG \cite{silver2014deterministic} and DQN \cite{mnih2013playing}. The LB-Aug is conducted as LB-Aug$_E$ for simplification. As shown in Table \ref{reinforcement}, Policy Gradient, DPG and AC has comparable performance, while DQN acts worse on all three evaluation metrics. In summary, directly predicting the policies (the former three) instead of choosing one policy according to the overall rewards (DQN) is more effective to the task searching label-based augmentation policies.

\subsubsection{Number of Folds}
\label{sec:number_of_folds}
\begin{figure}[!ht]
  \centering
  \subfigure[mA on Peta] 
  {\includegraphics[width=0.45\columnwidth]{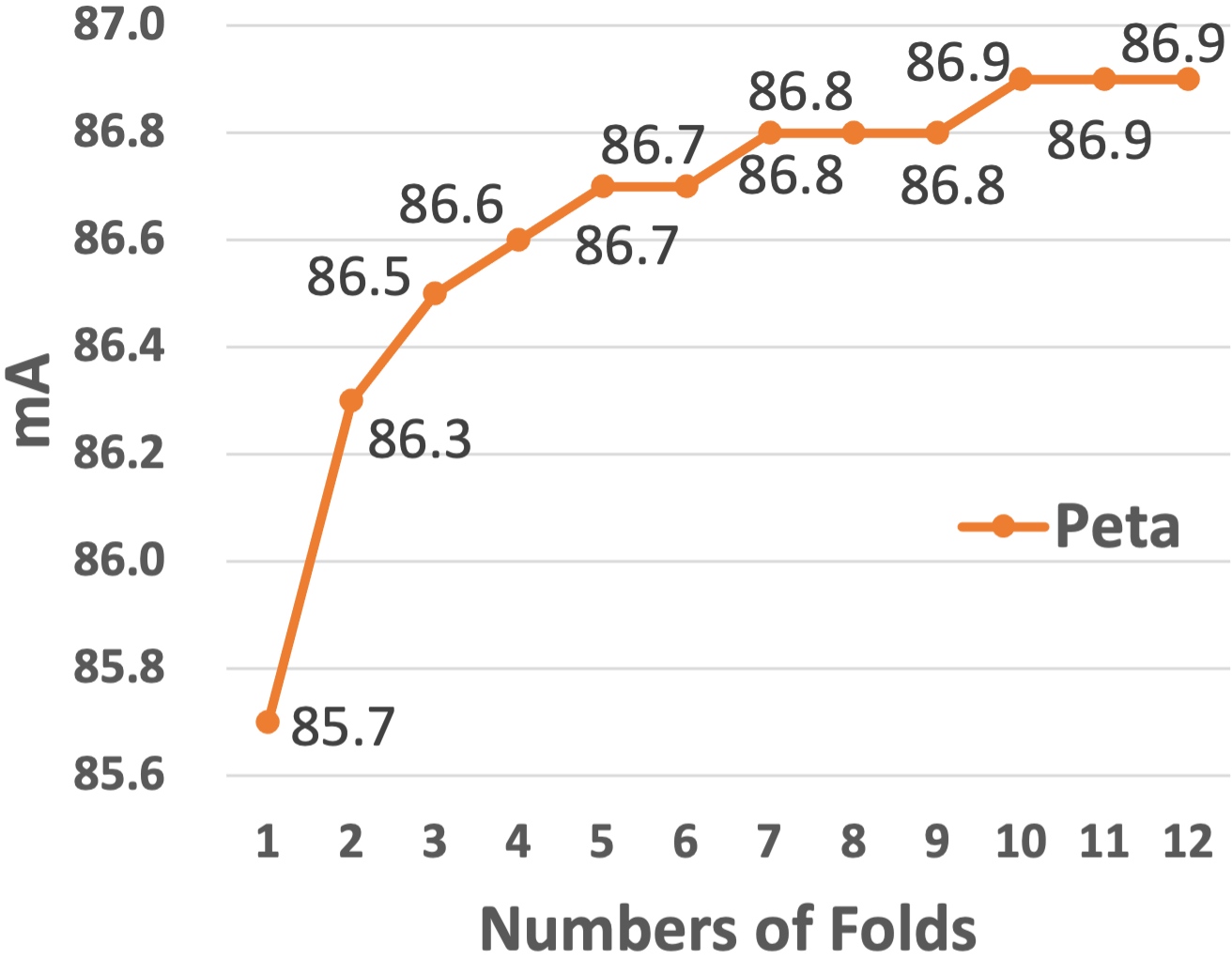} }\quad\ 
  \subfigure[mAP on MS-COCO]
  {\includegraphics[width=0.45\columnwidth]{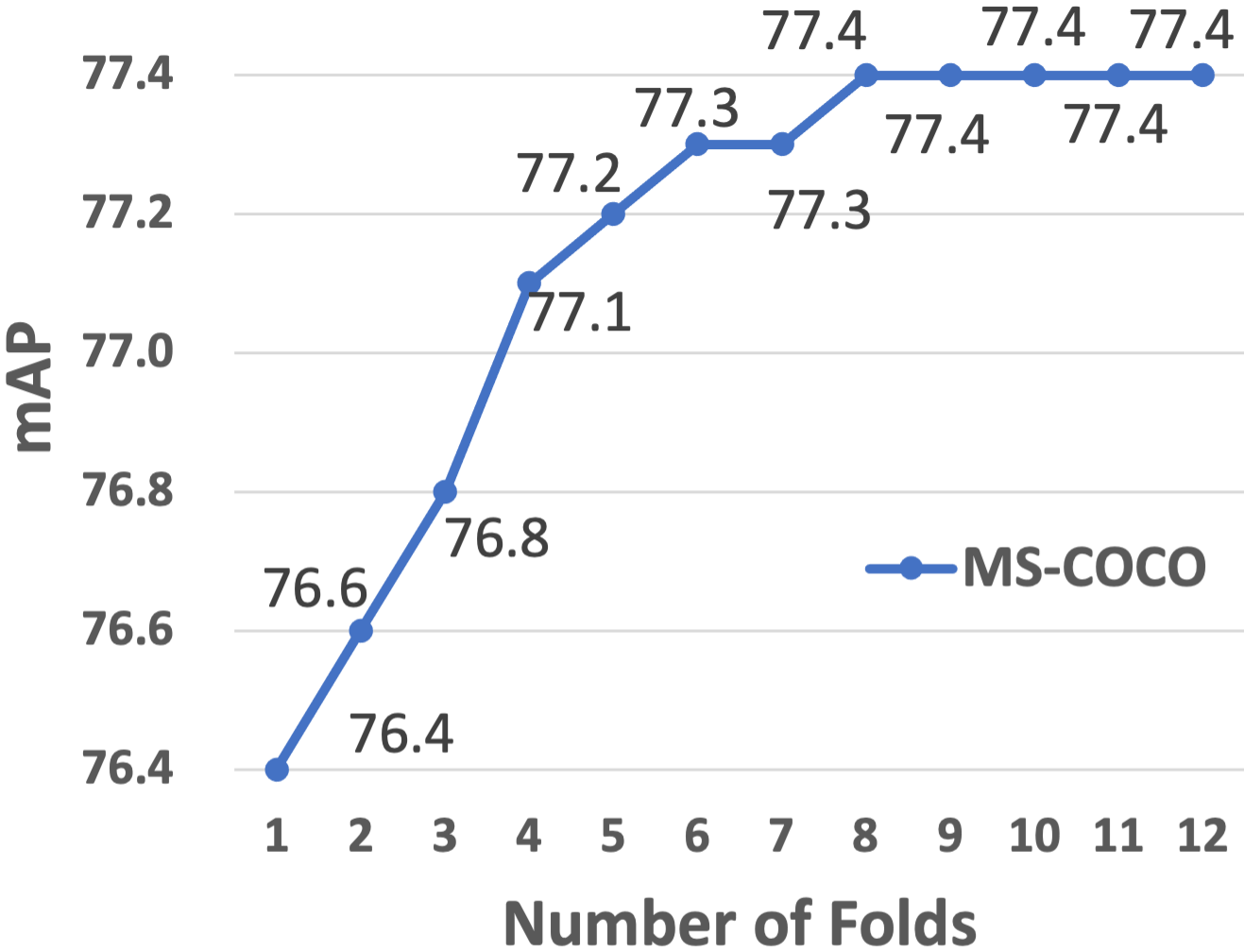} }
	\caption{Performance curve on fold numbers $K$. }
	\label{folds}
\end{figure}

We conduct K-fold stratified shuffling to split the origin training dataset into two segmentations, $D_M$ for training critic networks and $D_A$ to train the actor network. Finally, with label-based augmentation policies predicted by the actor network, a new multi-label classifier which has the same frameworks as critic but different weight parameters will be trained. As the actor network is trained on the union of $K$ splitted $D_A$ (Eqn. \ref{unions}), the number $K$ determines the scale of training set $\overline{D_A}$ of the actor. Figure \ref{folds} illustrates the results on Peta and MS-COCO of $K$ from 1 to 12. 

In total, the performance is monotonically improved when we enlarge the numbers of folds. The curves in Figure \ref{folds} indicates two terms of gains: 1) The case where the scale of $K$ is relatively small (less than 6 in Figure \ref{folds}). Noting that the $D_A$ is segmented with 1/5 samples of original training set $D_{train}$. In this case, bigger $K$ brings more training samples for the actor network, promoting better generalization. 2) The case where the scale of $K$ is relatively large (larger than 6 in Figure \ref{folds}). In this case, bigger $K$ no longer produces larger training set because $\overline{D_A} \approx D_{train}$. Nevertheless, the instances in $\overline{D_A}$ tend to have more corresponding critic networks, such that the reward of a policy on these instances could be more solid for the actor network training. In summary, using additional folds could monotonically boost the performance.

\section{Conclusion}

In this paper, we propose a novel label-based autoaugmentation method for multi-label image/video classification tasks. Benefitting from data augmentation policies fine-grained by considering instance labels, our method outperforms existing SOTA augmentation approaches by large margins in Peta, MS-COCO and Charades benchmarks. Extensive ablation study shows that the proposed LB-Aug method is compatible to most widely used architectures, with all of which the LB-Aug achieves the SOTA performance. Moreover, the method can be further extended to other multi-label style tasks such as ReID, Semantic Segmentation, and even unsupervised learning tasks wherever fine-grained information and rewards for actor network learning are available.

% \section*{Acknowledgments}

% \appendix
\bibliographystyle{unsrt}
\bibliography{main}

\newpage
\renewcommand{\appendixpagename}{Supplementary Materials}
\begin{appendices}

\section{Supplementary of Figure 1(a)}

\begin{figure}[!ht]
	\centering
	\includegraphics[width=0.95\columnwidth]{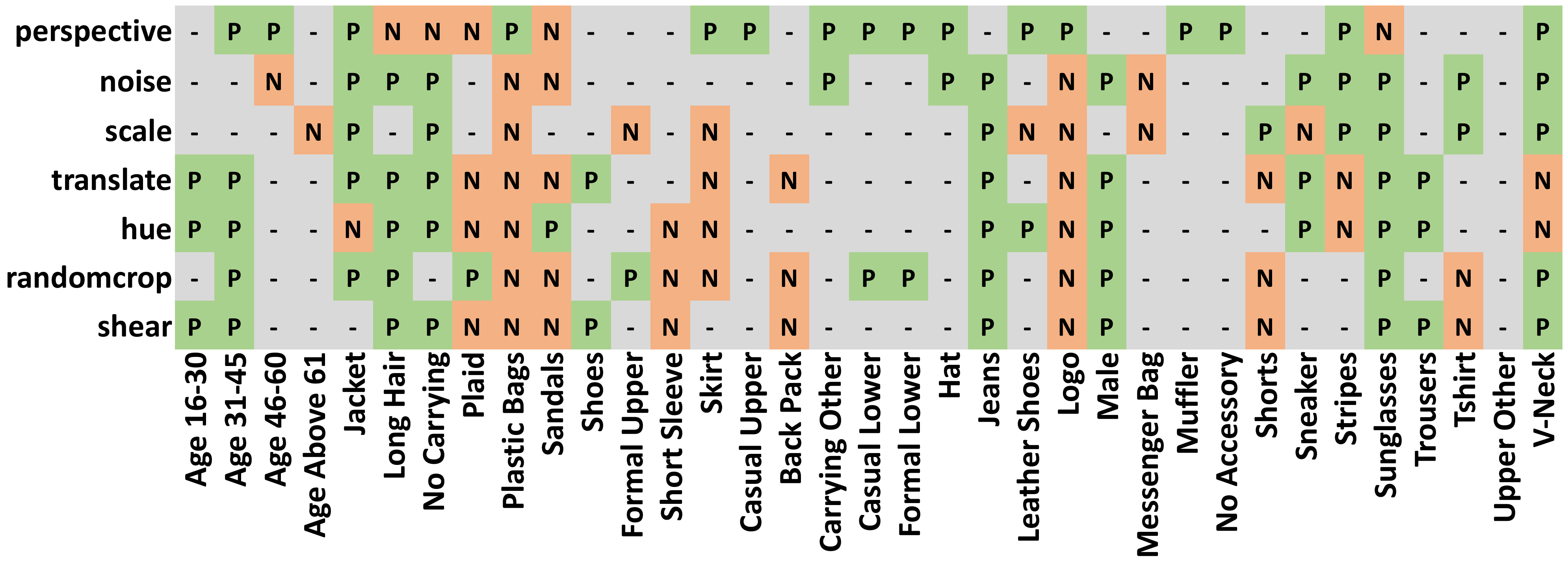}
	\caption{The gain/drop of augmentations on all labels of Peta. ``P'' indicates positive effects on the classification. ``N'' presents the effects are negative and ``-'' means inconspicuous influences.}
	\label{fig:influence_complete}
\end{figure}

\section{Depth of Actor Network}

\begin{figure}[!hb]
    \centering
    \subfigure[mA on Peta] 
    {\includegraphics[width=0.42\columnwidth]{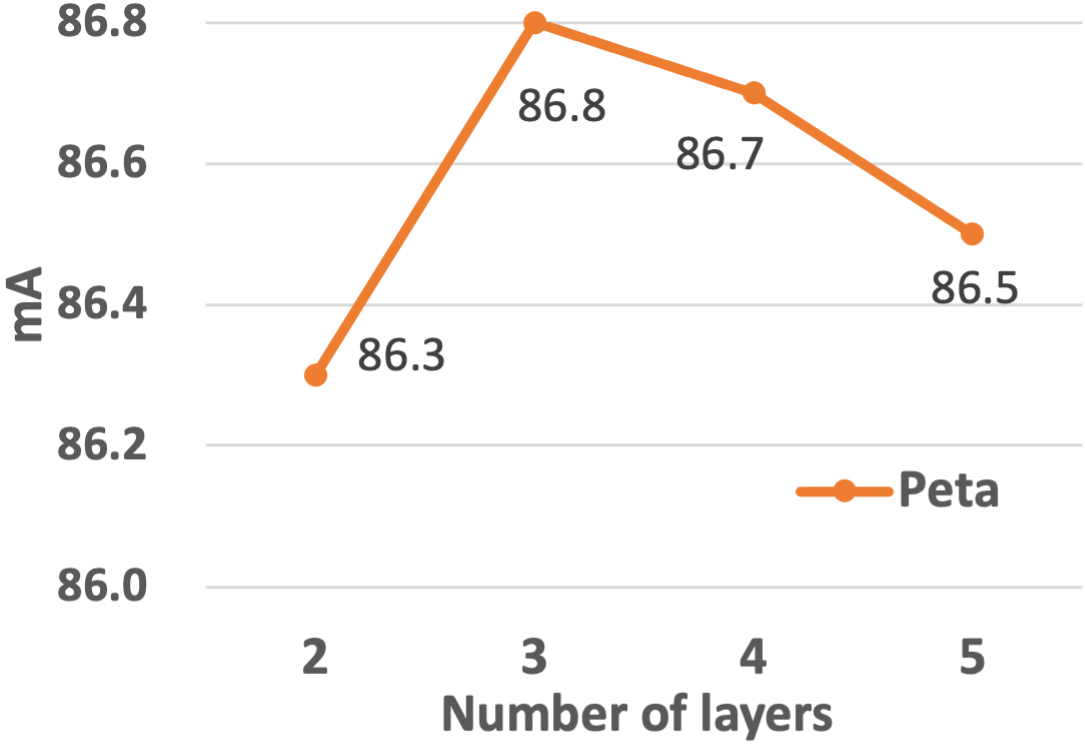}}\qquad\qquad
    \subfigure[mAP on MS-COCO]
    {\includegraphics[width=0.42\columnwidth]{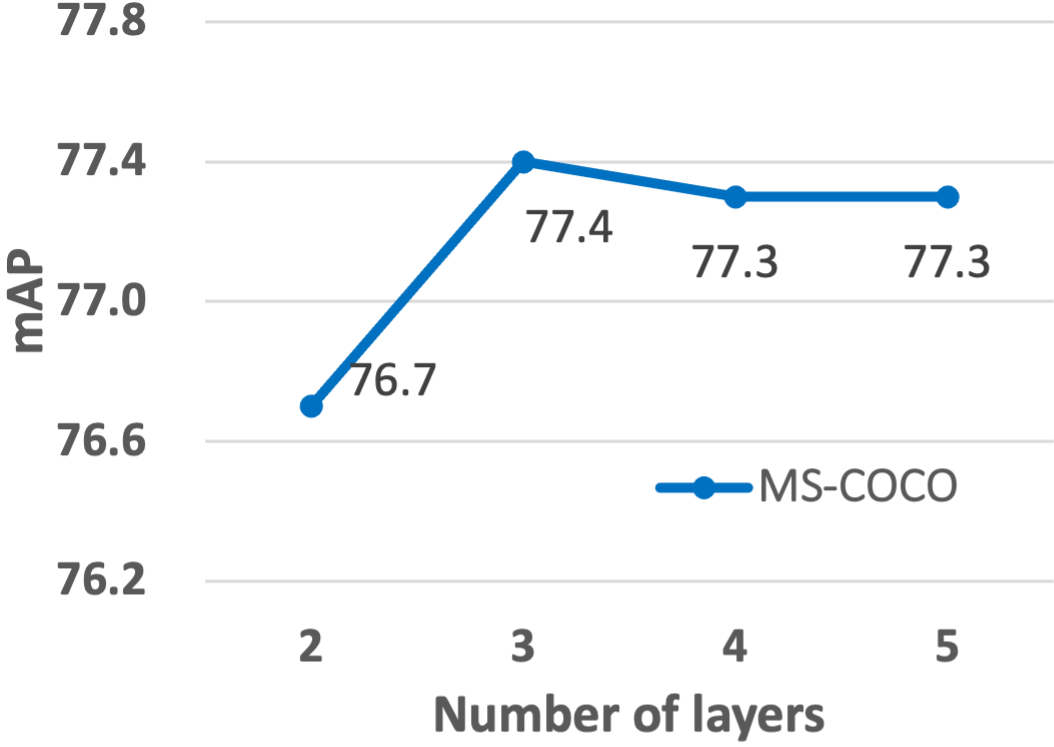}}
      \caption{The mA results on Peta and mAP results on MS-COCO. Each fully connected layer is followed by an ReLU and Dropout whose drop rate is 0.5.}
      \label{layers}
  \end{figure}

The actor network of our LB-Aug is constructed with fully connected layers which are equipped with ReLU to make a non-linear transformation and Dropout to overcome the overfitting problem. 
Figure \ref{layers} shows how the depth of the actor network (varying from 2 to 5) influences its performance. We can see that the depth of three works significantly better than depth 2, while as the network goes deeper, its performance on MS-COCO decreases slightly, while drops more distinctly on Peta. 
We think the reason why the depth of 2 has worse performance than that of the depth of 3 is the limitation of network capacity. Meanwhile, the performance decreases of those networks deeper than 3 are well possible caused by the ``overfitting'' problem, as the data augmentation policy learning task whose input is a multi-hot vector (whose dimension equals to the number of labels: 35 on Peta and 80 on MS-COCO) and output is the probabilities of taking each augmentation operation (whose dimension equals to the number of possible data augmentation operations: 16 in this paper), are relatively simpler than those complicated classification tasks such as the multi-label benchmarks used in this paper.
% different labels is a relatively simple classification
% we think is the task of finding augmentation policy for each label is much more simple compared with common deep learning tasks. 
% As the input of actor networks is a multi-hot vector whose length equals to the number of labels and the length of output is determined by the number of data augmentation operations (16 in this paper), the task of searching optimal policies for different labels is a relatively simple classification, and therefore suffers from ``overfitting'' problem when using a ``too deep'' actor network.

\end{appendices}

\end{document}